\documentclass[10pt,twocolumn,letterpaper]{article}

\usepackage{cvpr}              

\usepackage{graphicx}
\usepackage{amsmath}
\usepackage{amssymb}
\usepackage{booktabs}

\usepackage[pagebackref,breaklinks,colorlinks]{hyperref}

\usepackage[capitalize]{cleveref}
\crefname{section}{Sec.}{Secs.}
\Crefname{section}{Section}{Sections}
\Crefname{table}{Table}{Tables}
\crefname{table}{Tab.}{Tabs.}

\def\cvprPaperID{1****} 
\def\confName{CVPR}
\def\confYear{2022}

\begin{document}

\title{Time Gated Convolutional Neural Networks for Crop Classification}

\author{Longlong Weng\\
STAR.VISION\\
{\tt\small weng.longlong@star.vision}
\and
Yashu Kang\\
STAR.VISION\\
{\tt\small kang.yashu@star.vision}
\and
Kezhao Jiang\\
Columbia University\\
{\tt\small kj2589@columbia.edu}
\and
Chunlei Chen\\
STAR.VISION\\
{\tt\small chen.chunlei@star.vision}
}
\maketitle

\begin{abstract}
   This paper presented a state-of-the-art framework, Time Gated Convolutional Neural Network (TGCNN) that takes advantage of temporal information and gating mechanisms for the crop classification problem. Besides, several vegetation indices were constructed to expand dimensions of input data to take advantage of spectral information. Both spatial (channel-wise) and temporal (step-wise) correlation are considered in TGCNN. Specifically, our preliminary analysis indicates that step-wise information is of greater importance in this data set. Lastly, the gating mechanism helps capture high-order relationship. Our TGCNN solution achieves $0.973$ F1 score, $0.977$ AUC ROC and $0.948$ IoU, respectively. In addition, it outperforms three other benchmarks in different local tasks (Kenya, Brazil and Togo). Overall, our experiments demonstrate that TGCNN is advantageous in this earth observation time series classification task.
\end{abstract}

\section{Introduction}
\label{sec:intro}

Satellite earth observation (EO) data has been widely used in various domains including urban management, climate change, agriculture, environmental monitoring \cite{singh2017,chuvieco2010,yi2020,anderson2017}, etc. The recent advances in both EO technology and artificial intelligence community have gathered attentions in applying machine learning techniques to EO data with spatial, spectral and temporal characteristics. One of the most prominent machine learning approaches is supervised machine learning. As often noted by researchers, the availability of high-quality data is of great importance to the performance of a classification system. It is critical to have large number of training data to establish a deep learning model such as Convolutional Neural Networks (CNN) suitable for global generalization. 

When it comes to time series problems, the EO data usually has two spatial dimensions (width and height of channels), one spectral dimension (number of bands) and temporal dimension (\eg time step). Existing studies proposed various deep learning frameworks to solve these challenges \cite{kidger2020,liu2021gated,oguiza2020tsai,fauvel2021xcm,liu2021pay}, but the approach of dealing with such complex space into a feature space with relevant information is crucial. For instance, Tseng et al. took the advantage of meta learning and achieved satisfying results on local tasks while training with a global data set \cite{tseng2021cropharvest,tseng2021learning}. Model Agnostic Meta Learning (MAML) is compatible with any model trained with gradient descent and applicable to classification tasks \cite{finn2017model}. In our study, we also considered MAML as a benchmark for comparison.

On the CropHarvest challenge track, the released data is spatially and semantically comprehensive, consisting of more than $90,000$ samples with crop/non-crop and agricultural class labels collected from various sources covering $343$ labels \cite{tseng2021cropharvest}. The input variables cover multiple modes containing Sentinel-1 synthetic aperture radar information, Sentinel-2 multispectral information, ERA5 climatological variables, SRTM DEM topographic variables, \etc. Such a unique and global data set offers new opportunities for research and a variety of applications as well as new challenges.


\section{Methodology}
\label{sec:methodology}
\begin{table*}
  \centering
  \begin{tabular}{@{}llc@{}}
    \toprule
    Vegetation index & Abbreviation & Calculation \\
    \midrule
    Normalized Difference Vegetation Index & NDVI & $\frac{NIR-R}{NIR+R}$\\
    \midrule
    Soil Adjusted Vegetation Index & SAVI & $\frac{NIR-R}{NIR+R+L}\times(1+L)$\\
     \midrule
    Simple Ratio Index & SR & $\frac{NIR}{R}$\\
     \midrule
    Red-Edge Chlorophyll Index (RECI) & RECI & $\frac{NIR}{R}-1$\\
     \midrule
    Normalized Difference Red Edge Index & NDRE & $\frac{NIR-R}{NIR+R}$\\
     \midrule
    Modified Soil Adjusted Vegetation Index & MSAVI & $\frac{2\times NIR+1-\sqrt{(2\times NIR+1)^2 - 8\times(NIR-R)}}{2}$\\
     \midrule
    Normalized Difference Water Index & NDWI & $\frac{G-NIR}{NIR+G}$\\
     \midrule
    Green Chlorophyll Index & GCI & $\frac{NIR}{G} - 1$\\
    \bottomrule
  \end{tabular}
  \centering
  \caption{Constructed vegetation indices}
  \label{tab:one}
\end{table*}

\subsection{Feature engineering}

In the CropHarvest data set, one of the input variables is normalized difference vegetation index (NDVI), calculated by using NIR (near-infrared, wavelength $835$ nm) and R (Red, wavelength $665$ nm) bands, because vegetation generally absorbs red and reflects NIR spectral information \cite{tseng2021cropharvest}. Among various types of vegetation indices, NDVI is among the most frequently used. However, in high vegetation covered areas, NDVI may be saturated, in addition to its non-linear relationship with bio-physiological variations. Inclusion of more vegetation indices could be beneficial. Thus, we constructed several more indices as additions to the input variables that also involve G (green, wavelength $560$ nm) and RE (red edge, wavelength $740$ nm) bands. These vegetation indices are soil adjusted vegetation index (SAVI), simple ratio index (SR), red edge chlorophyll (RECI), normalized difference red edge index (NDRE), modified soil adjusted vegetation index (MSAVI), normalized difference water index (NDWI) and green chlorophyll index (GCI) \cite{xue2017significant,rouse1973monitoring,chen1996evaluation,gitelson2003relationships,thompson2019using,gao1996ndwi}, as shown in \cref{tab:one}.

\begin{figure*}
  \includegraphics[width=1\linewidth]{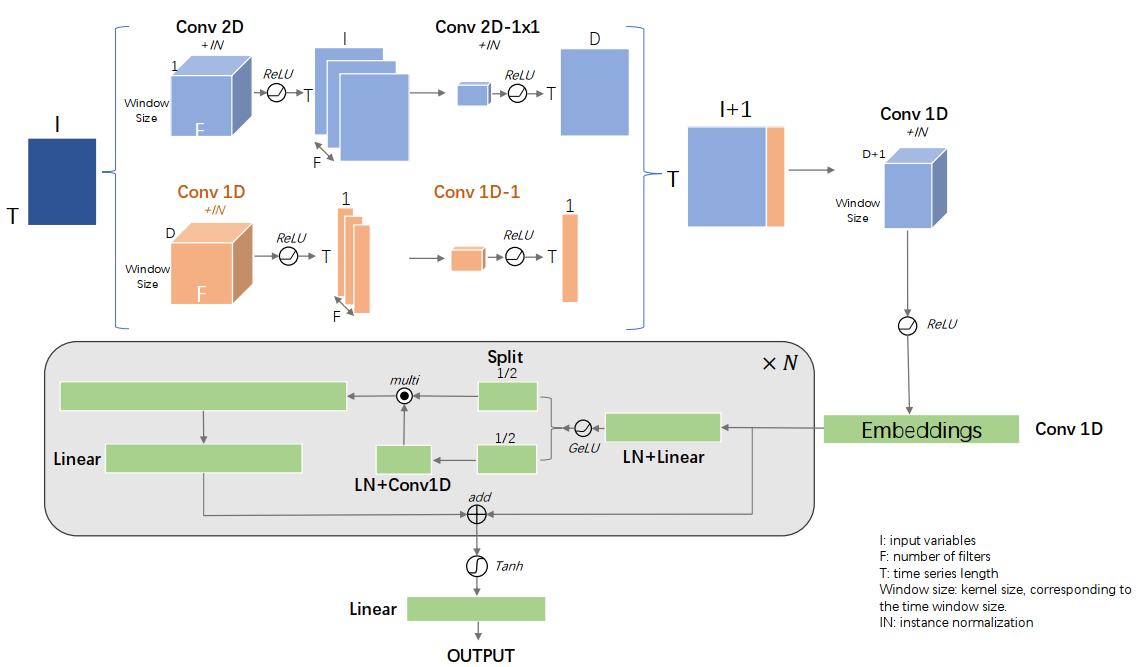}
  \caption{Overview of the TGCNN architecture}
  \label{fig:one}
\end{figure*}

\subsection{Architecture overview of TGCNN}

Here we propose a novel Time Gated Convolutional Neural Network (TGCNN) framework for time series classification, as illustrated in \cref{fig:one}. The framework incorporates the gating mechanism while temporal information is processed and converted into a different feature map, hence the name TGCNN. Firstly, most input variables of the input data are fed into 2D convolutional filters while temporal information goes through 1D convolutional filters, which will yield two feature maps. This guarantees that temporal information is fully integrated to discriminate between the target classes. 

In \cref{fig:one}, the upper light blue section indicates processing explanatory variables through a 2D convolutional filter, converted to feature map to reduce dimensionality with a $1\times1$ convolutional filter. The convolutional block consists of a 2D convolutional layer, an instance normalization layer and a ReLU activation layer. Instance normalization proves to be beneficial to improve training convergence and reduce the sensitivity to network hyperparameters as a mini-batch of data is normalized for each observation independently. The ReLU activation layer provides the nonlinearity here. Window size refers to the time window size acting as a hyperparameter, i.e. window size x1 means each explanatory variable is considered to extract discriminative features. Similarly, the middle section mirrors the explanatory variables except that for temporal information, 2D convolutional layers are replaced with 1D ones. Then the spatial (channel-wise) and temporal (step-wise) feature maps are concatenated, processed with another 1D convolutional filter and processed with the next gating module. 

In recent years, Transformer has emerged as state-of-the-arts in various domains, including time-series challenges \cite{liu2021gated,oguiza2020tsai}. Along with the success of Transformer, some studies sought further development in CNN and proposed to utilize tokenization and embedding process at the stem as an alternative of convolution \cite{fauvel2021xcm,liu2021pay}. Inspired by these works, the lower gating module in \cref{fig:one} consists of N blocks with identical structures, each resembling a Gated Linear Units (GLU). We find it efficient to adopt this self-attention-like gating structure for its simplicity. Its computation cost is linear over the input channel size and quadratic over the time series length. Note that after the GeLU activation layer, the features are split into two parts along the channel dimension to take advantage of gating and multiplicative function.

\subsection{Performance tricks}

For the final activation layer, the tanh nonlinearity restricts the rate of change of the hidden state in order to avoid large initial losses, which proves to be beneficial in this gating module. This may be similar to RNNs where the amount of change between time steps are constrained.

Since the discriminant information such as spatial (channel-wise) and temporal (step-wise) variables are crucial in multivariate time series problems, we performed analysis on explicitly extracting spatial features alone and temporal features alone. The step-wise features indicated higher importance. This also echoes the benefits of separately processing variables related to temporal information.  

\section{Analysis Results}
\label{sec:analysis}

The number of dimensions of original data set has expanded with additional vegetation indices corresponding to \cref{tab:one}. These feature engineering were mainly constructed using Sentinel-2 multispectral information and applied to both training and test data. We assessed the performance of our model by comparing with Gated Transformer \cite{liu2021gated}, and benchmarks from \cite{tseng2021cropharvest} with two different setups (Random weights and MAML initiation, respectively). 

\begin{figure}[t]
  \centering
  \includegraphics[width=1\linewidth]{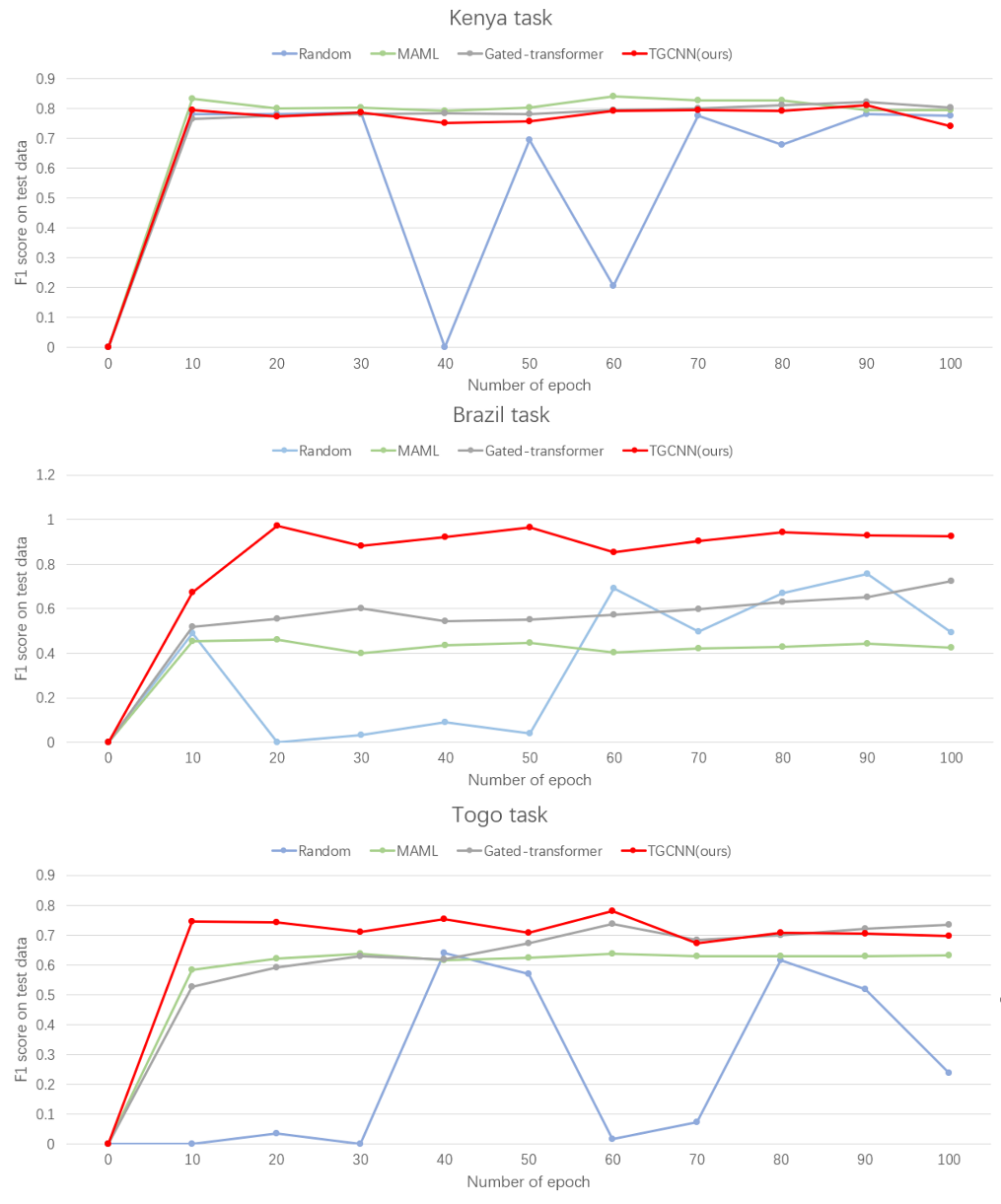}

   \caption{Model assessment with different epoch numbers}
   \label{fig:two}
\end{figure}

As \cref{fig:two} demonstrates, it only takes approximately $20$ epochs for our TGCNN model to ascend to a desired level for the Brazil task, significantly outperforming the other three models. As for the Kenya and Togo tasks, all models are showing comparable performances. In addition, F1 score performances on CropHarvest test data of the four different model settings are evaluated and presented in \cref{tab:two}. One can conclude that our model outperformed all other benchmarks in different local tasks although the advantage is narrower for the Brazil task. 

\begin{table}[t]
  \centering
  \begin{tabular}{@{}lll@{}}
    \toprule
    Tasks & Model & F1-score \\
    \midrule
    Brazil & \begin{tabular}{@{}l@{}}Random \\ MAML \\ Gated transformer \\ TGCNN (ours)\end{tabular} &
    \begin{tabular}{@{}l@{}}$0.9209$ \\ $0.9477$ \\ $0.962$ \\ $\mathbf{0.9739}$ \end{tabular}\\
    \midrule
    Kenya & \begin{tabular}{@{}l@{}}Random \\ MAML \\ Gated transformer \\ TGCNN (ours)\end{tabular} & \begin{tabular}{@{}l@{}}$0.7812$ \\ $0.8315$ \\ $0.8056$ \\ $\mathbf{0.8543}$ \end{tabular}\\
     \midrule
    Togo & \begin{tabular}{@{}l@{}}Random \\ MAML \\ Gated transformer \\ TGCNN (ours)\end{tabular} & \begin{tabular}{@{}l@{}}$0.6736$ \\ $0.6569$ \\ $0.7644$ \\ $\mathbf{0.7948}$ \end{tabular}\\
    \bottomrule
  \end{tabular}
  \caption{Model performance comparisons on different tasks}
  \label{tab:two}
\end{table}

For different tasks, the data might show leaning towards step-wise information or channel-wise information, especially for challenges in another domain. The assumption is that for a crop classification problem, temporal information should be crucial differentiator. Our preliminary results on the Brazil task provided validation to this assumption when we obtained a better performance (F1 score $0.94$) with step-wise only model than channel-wise model (F1 score $0.70$). Therefore, by incorporating this mechanism in a global data set, TGCNN yields better classification results than the other benchmarks overall. However, the imbalance issue in the data set still poses a challenge to the solution when it comes to multiple regional data subsets. The modeling results are heavily impacted by regions that have significantly more samples. 

\section{Concluding Remarks}
\label{sec:conclude}

In this paper, we presented a state-of-the-art framework, Time Gated Convolutional Neural Network (TGCNN). The model takes advantage of temporal information and gating mechanisms for the crop classification problem. More than $90,000$ geographically diverse samples formed a challenging task that requires analysis on multimodal data. TGCNN can allow modeling both spatial and temporal correlations effectively utilized convolutional neural network and gating mechanism. The results also paved the way for more in-depth study of machine learning techniques on earth observation time series data in future research and applications.

{\small
\bibliographystyle{ieee_fullname}
\bibliography{PaperForReview}
}

\end{document}